\documentclass[11pt]{article}
\usepackage{acl2010}
\usepackage{times}
\usepackage{url}
\usepackage{latexsym}
\usepackage{amsmath}
\usepackage{multirow}
\usepackage{arydshln}
\usepackage{graphicx}
\usepackage{tikz}
\usepackage{tikz-cd}
\usetikzlibrary{matrix,calc,arrows,arrows.meta}

\title{Commonsense Reasoning \\ Using \WORDNET{} and \SUMO{}: a Detailed Analysis} 
\author{}

\author{Javier \'{A}lvez\\
  LoRea Group\\
  \\
  \And
  Itziar Gonzalez-Dios\\
  Ixa Group\\
  University of the Basque Country (UPV/EHU)\\
  {\tt \{javier.alvez,itziar.gonzalezd,german.rigau\}@ehu.eus} \And
  German Rigau\\
  Ixa Group\\
  }
\date{}

\newcommand{\WORDNET}{{WordNet}}
\newcommand{\SUMO}{SUMO}

\newcommand{\ADIMENSUMO}{Adimen-SUMO}
\newcommand{\FOLSUMO}{FOL-SUMO}

\newcommand{\textConstant}[1]{{\it{#1}}}

\newcommand{\textPredicate}[1]{{\it{#1}}}

\newcommand{\SUMOInstanceSymbol}{o}
\newcommand{\SUMOClassSymbol}{c}
\newcommand{\SUMOIndividualRelationSymbol}{r}
\newcommand{\SUMOIndividualAttributeSymbol}{a}

\newcommand{\SUMOInstance}[1]{{\textConstant{#1}$_\SUMOInstanceSymbol$}}
\newcommand{\SUMOClass}[1]{{\textConstant{#1}$_\SUMOClassSymbol$}}
\newcommand{\SUMOIndividualRelation}[1]{{\textConstant{#1}$_\SUMOIndividualRelationSymbol$}}

\newcommand{\SUMOIndividualAttribute}[1]{{\textConstant{#1}$_\SUMOIndividualAttributeSymbol$}}

\newcommand{\SUMOClassTikZ}[1]{{\constant{#1}_\SUMOClassSymbol}}

\newcommand{\synset}[3]{{\it{#1}$_{#3}^{#2}$}}
\newcommand{\pair}[3]{{\it{#1}(#2,#3)}}

\newcommand{\equivalenceMappingSymbol}{=}
\newcommand{\subsumptionMappingSymbol}{+}
\newcommand{\instantiationMappingSymbol}{@}

\newcommand{\equivalenceMapping}[1]{{\textConstant{#1}$\equivalenceMappingSymbol$}}
\newcommand{\subsumptionMapping}[1]{{\textConstant{#1}$\subsumptionMappingSymbol$}}
\newcommand{\instantiationMapping}[1]{{\textConstant{#1}$\instantiationMappingSymbol$}}

\newcommand{\equivalenceMappingRelation}{{\it{equivalence}}}
\newcommand{\subsumptionMappingRelation}{{\it{subsumption}}}
\newcommand{\instantiationMappingRelation}{{\it{instantiation}}}

\newcommand{\synsetTikZ}[3]{{\it{#1}_{#3}^{#2}}}

\newcommand{\equivalenceMappingTikZOfConcept}[1]{{#1}\hspace{-4pt}\equivalenceMappingSymbol}
\newcommand{\subsumptionMappingTikZOfConcept}[1]{{#1}\subsumptionMappingSymbol}

\newcommand{\minitab}{\hspace{10pt}}
\newcommand{\tab}{\minitab\minitab}
\newcommand{\doubletab}{\tab\tab}
\newcommand{\connective}[1]{\bf #1 \;}
\newcommand{\predicate}[1]{\rm #1}
\newcommand{\constant}[1]{\rm #1}
\newcommand{\variable}[1]{\tt ?#1}

\begin{document}

\maketitle

\begin{abstract} % no longer than 200 words.
We describe a detailed analysis of a sample of large benchmark of commonsense reasoning problems that has been automatically obtained from \WORDNET{}, \SUMO{} and their mapping. The objective is to provide a better assessment of the quality of both the benchmark and the involved knowledge resources for advanced commonsense reasoning tasks. By means of this analysis, we are able to detect some knowledge misalignments, mapping errors and lack of knowledge and resources. Our final objective is the extraction of some guidelines towards a better exploitation of this commonsense knowledge framework by the improvement of the included resources.
%by means of the detection of possible knowledge misalignment and discrepancies between the knowledge of \WORDNET{}, \SUMO{} and their mapping.
\end{abstract}

\section{Introduction} \label{section:Introduction}

%Recently, Artificial Intelligence has shown great advances in multiple research areas, but there is one critical area where limited progress has been shown: commonsense knowledge representation and commonsense reasoning \cite{mccarthy1989artificial,BBK01,BlB05,minsky2007emotion,DaM15}. The work introduced in this paper proposes to advance a step forward in this research line by providing a new black-box evaluation framework of first-order logic (FOL) \SUMO{}-based ontologies \cite{Niles+Pease'01} that exploits the world knowledge from \WORDNET{} \cite{Fellbaum'98} and its mapping into \SUMO{} \cite{Niles+Pease'03}.

Any ontology tries to provide an explicit formal semantic specification of the concepts and relations in a domain \cite{noy2001ontology,GuW02,GuW04,Gru09,StS09,ALR12}. As with other software artefacts, ontologies typically have to fulfil some previously specified requirements. Usually both the creation of ontologies and the verification of its requirements are manual tasks that require a significant amount of human effort. In the literature, some methodologies collect the experience in ontology development \cite{GFC04,GuW04} and in ontology verification \cite{GCC06}. 

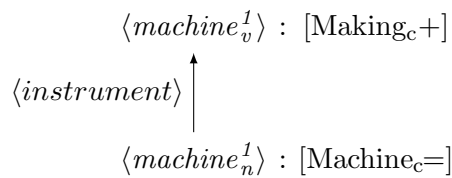
\begin{figure}[ht]
\centering
\begin{tikzpicture}[>=triangle 60]
\matrix[matrix of math nodes,column sep={-3pt},row sep={50pt,between origins},nodes={asymmetrical rectangle}] (s)
{
|[name=instrument]| \langle \synsetTikZ{machine}{1}{v} \rangle & : & |[name=instrumentMappingClass]| [ \subsumptionMappingTikZOfConcept{\SUMOClassTikZ{Making}} ] \\
|[name=process]| \langle \synsetTikZ{machine}{1}{n} \rangle & : & |[name=processMappingClass]| [ \equivalenceMappingTikZOfConcept{\SUMOClassTikZ{Machine}} ] \\
};
\draw[-latex] (process) -- node[left] {\(\langle instrument \rangle\)} (instrument);
\end{tikzpicture}\caption{An example of \WORDNET{} and its mapping to \SUMO{}}
\label{fig:introduction}
\end{figure}

%In \cite{ALR19}, the authors propose a method for the semi-automatic creation of CQs that enables the evaluation of the {\it competency} of \SUMO{}-based ontologies in the sense proposed in \cite{GrF95}.

In order to evaluate the {\it competency} of \SUMO{}-based ontologies in the sense proposed by Gr{\"u}ninger and Fox \shortcite{GrF95}, \'Alvez et al. \shortcite{ALR19} propose a method for the semi-automatic creation of {\it competency questions} (CQs). Concretely, they adapt the methodology to evaluate the ontologies so that it can be automatically applied using automated theorem provers (ATPs). The construction of CQs is based on several predefined {\it question patterns} (QPs) that yield a large set of problems (dual conjectures) by using information from \WORDNET{} and its mapping into \SUMO{}. For example, the synsets \synset{machine}{1}{v} and \synset{machine}{1}{n} are related by the morphosemantic relation \textPredicate{instrument} in the {\it Morphosemantic Links} \cite{FOC09} of \WORDNET{}, as depicted in Figure \ref{fig:introduction}. In the same figure, the mappings of the synsets are also provided: \synset{machine}{1}{n} and \synset{machine}{1}{v} are connected to \equivalenceMapping{\SUMOClass{Machine}} and \subsumptionMapping{\SUMOClass{Making}}, where the symbol `$=$' means that \synset{machine}{1}{n} is semantically equivalent to the \SUMOClass{Machine}, while `$+$' means that the semantics of \SUMOClass{Making} is more general than the semantics of \synset{machine}{1}{v}. Hence, it is possible to state the the relationship of \synset{machine}{1}{n} and \synset{machine}{1}{v} in terms of \SUMO{} as follows:
\begin{flalign}
% (forall (?Y)
% 	(=> 
% 		($instance ?Y Machine)
% 		(exists (?X)
% 			(and 
% 				($instance ?X Making)
% 				(instrument ?X ?Y)
% 			)
% 		)
% 	)
% )
\minitab & ( \connective{forall} \; ( \variable{Y} ) & \label{CQ:machine} \\
 & \tab ( \connective{=>} & \nonumber \\
 & \tab \tab ( \predicate{instance} \; \variable{Y} \; \constant{Machine} ) & \nonumber \\
 & \tab \tab ( \connective{exists} \; ( \variable{X} ) & \nonumber \\
 & \tab \tab \tab ( \connective{and} & \nonumber \\
 & \tab \tab \tab \tab ( \predicate{instance} \; \variable{X} \; \constant{Making} ) & \nonumber \\
 & \tab \tab \tab \tab ( \predicate{instrument} \; \variable{X} \; \variable{Y} ) ) ) ) ) & \nonumber
\end{flalign}
The problem that results from Figure \ref{fig:introduction} consists of the above conjecture, which is considered to be true according to our commonsense knowledge, and its negation, which is therefore assumed to be false.

State-of-the-art ATPs for first-order logic (FOL) such as Vampire \cite{KoV13} or E \cite{Sch02} have been proved to provide advanced reasoning support to large FOL conversions of expressive ontologies \cite{RRG05,HoV06,PeS07,ALR12}. However, the semi-decidability of FOL and the poor scalability of the known decision procedures have been usually identified as the main drawbacks for the practical use of FOL ontologies. In particular, given an unsolved problem (i.e. a problem such that ATPs do not find any proof for its pair of conjectures) it is not easy to know if (a) the conjectures are not entailed by the ontology or (b) although some of the conjecture is entailed, ATPs have not been able to find the proof within the provided execution-time and memory limits. On the contrary, given a solved problem, it is hard to know whether the solution is obtained for a good reason, because an expected result does not always indicate a correct ontological modelling.

In this paper, we provide a detailed analysis of the large commonsense reasoning benchmarks created semi-automatically by \cite{ALR17,ALR19}. The aim of this analysis is to shed light on the commonsense reasoning capabilities of both the benchmark and the involved knowledge resources. To that end, we have randomly selected a sample of 169 problems (1\% of the total) following a uniform distribution and manually inspected their source knowledge and results. By means of this detailed analysis, we are able to evaluate the quality of automatically created benchmarks of problems and to detect hidden problems and misalignments between the knowledge of \WORDNET{}, \SUMO{} and their mapping.

{\it Outline of the paper}. In order to make the paper self-contained, we first introduce the ontology, the mapping to \WORDNET{} and the evaluation framework in Section \ref{section:Evaluation}. Next, we provide a full summary and the main conclusions obtained from our manual analysis in Section \ref{section:Analysis}. Then, we individually examine some of the problems in Section \ref{section:SomeProblems}. Finally, we provide some conclusions and discuss future work in Section \ref{section:Conclusion}.

\section{Commonsense Reasoning Framework} \label{section:Evaluation}

In this section we describe briefly the whole commonsense reasoning reasoning framework. First, we present the knowledge resources needed and then the reasoning framework.

\subsection{Resources} \label{section:resources}

The resources we present in this section are \FOLSUMO{}, \WORDNET{} and the semantic mapping between them.

%\section{\FOLSUMO{}} \label{section:SUMO}

\SUMO{}\footnote{\url{http://www.ontologyportal.org}} \cite{Niles+Pease'01} is an upper level ontology proposed as a starter document by the IEEE Standard Upper Ontology Working Group. \SUMO{} is expressed in SUO-KIF (Standard Upper Ontology Knowledge Interchange Format \cite{Pea09}), which is a dialect of KIF (Knowledge Interchange Format \cite{Richard+'92}). The syntax of both KIF and SUO-KIF goes beyond FOL and, therefore, \SUMO{} axioms cannot be directly used by FOL ATPs without a suitable transformation \cite{ALR12}.

To the best of our knowledge, there are two main proposals for the translation of the two upper levels of \SUMO{} into a FOL formulae that are described in Pease and Sutcliffe, \shortcite{PeS07}, Pease et al. \shortcite{pease2010large} and \'Alvez et al. \shortcite{ALR12} respectively. Both proposals have been developed under the {\it Open World Assumption} (OWA) \cite{Rei78} and are currently included in the {\it Thousands of Problems for Theorem Provers} (TPTP) problem library\footnote{\url{http://www.tptp.org}} \cite{Sut09}. In this paper, we use \ADIMENSUMO{} v2.6, which is freely available at \url{https://adimen.si.ehu.es/web/AdimenSUMO}. From now on, we refer to \ADIMENSUMO{} v2.6 as \FOLSUMO{}.

The knowledge in \SUMO{}, and therefore in \FOLSUMO{}, is organised around the notions of {\it instance} and {\it class}. These concepts are respectively defined in \SUMO{} by means of the predicates \textPredicate{instance} and \textPredicate{subclass}.\footnote{It is worth noting that term {\it instance} is overloaded since it denotes both the \SUMO{} predicate and the \SUMO{} concepts that are defined by using that predicate.} Additionally, \SUMO{} also differentiates between {\it relations} and {\it attributes}, which are organized using the predicates \textPredicate{subrelation} and \textPredicate{subAttribute} respectively. For simplicity, from now on we denote the nature of \SUMO{} concepts by adding as subscript the symbols $\SUMOInstanceSymbol{}$ (\SUMO{} instances that are neither relations nor attributes), $\SUMOClassSymbol{}$ (\SUMO{} classes that are neither classes of relations nor classes of attributes), $\SUMOIndividualRelationSymbol{}$ (\SUMO{} relations) and $\SUMOIndividualAttributeSymbol{}$ (\SUMO{} attributes). For example: \SUMOInstance{Waist}, \SUMOClass{Artifact}, \SUMOIndividualRelation{customer} and \SUMOIndividualAttribute{Female}.

%\section{The Mapping between \WORDNET{} and \SUMO{}} \label{section:Mapping}

\WORDNET{} \cite{Fellbaum'98} is linked with \SUMO{} by means of the mapping described in Niles and Pease \shortcite{Niles+Pease'03}. This mapping connects \WORDNET{} synsets to terms in \SUMO{} using three relations: \equivalenceMappingRelation{}, \subsumptionMappingRelation{} and \instantiationMappingRelation{}. We denote the mapping relations by concatenating the symbols `$\equivalenceMappingSymbol$' (\equivalenceMappingRelation{}), `$\subsumptionMappingSymbol$' (\subsumptionMappingRelation) and `$\instantiationMappingSymbol$' (\instantiationMappingRelation) to the corresponding \SUMO{} concept. For example, the synsets \synset{horse}{1}{n}, \synset{education}{4}{n} and \synset{zero}{1}{a} are connected to \equivalenceMapping{\SUMOClass{Horse}}, \subsumptionMapping{\SUMOClass{EducationalProcess}} and \instantiationMapping{\SUMOClass{Integer}} respectively. \equivalenceMappingRelation{} denotes that the related \WORDNET{} synset and \SUMO{} concept are equivalent in meaning, whereas \subsumptionMappingRelation{} and \instantiationMappingRelation{} indicate that the semantics of the \WORDNET{} synset is less general than the semantics of the \SUMO{} concept. In particular, \instantiationMappingRelation{} is used when the semantics of the \WORDNET{} synsets refers to a particular member of the class to which the semantics of the \SUMO{} concept is referred. %From now on, we say that a \WORDNET{} synset is {\it less general} (or {\it more specific}) than the \SUMO{} concepts to which the synset is connected using \subsumptionMappingRelation{} or \instantiationMappingRelation{}, and {\it vice versa}.

The mapping between \WORDNET{} and \SUMO{} can be translated into the language of \SUMO{} by means of the proposal introduced in \'Alvez et al. \shortcite{ALR17}. This translation characterises the mapping information of a synset in terms of \SUMO{} instances by using equality (for \SUMO{} instances) and the \SUMO{} predicates \SUMOIndividualRelation{instance} and \SUMOIndividualRelation{attribute} (for \SUMO{} classes and attributes respectively). %One \SUMO{} statement is constructed for each synset by introducing new variables. The quantification of the introduced variables is determined by the mapping relation that is used for connecting the given synset. Next, we briefly describe the translation of the mapping information of synsets connected to a single \SUMO{} concept:
For example, the noun synsets \synset{smoking}{1}{n} and \synset{breathing}{1}{n} are respectively connected to \equivalenceMapping{\SUMOClass{Smoking}} and \equivalenceMapping{\SUMOClass{Breathing}}. Thus, the \SUMO{} statements that result by following the proposal described in \'Alvez et al. \shortcite{ALR17} is:
\begin{flalign}
\doubletab & ( \predicate{instance} \; \variable{X} \; \constant{Smoking} ) & \label{subCQ:smoking} \\
 & ( \predicate{instance} \; \variable{X} \; \constant{Breathing} ) & \label{subCQ:breathing}
\end{flalign}

\subsection{Evaluation Framework} \label{section:framework}

\begin{table}[!t]
\centering
\resizebox{\columnwidth}{!}{
\begin{tabular}{llr}
\hline
\multicolumn{1}{c}{{\bf \WORDNET{} Relation}} & \multicolumn{1}{c}{{\bf QP}} & \multicolumn{1}{c}{{\bf Problems}} \\
\hline
\multirow{4}{*}{{\it Hyponymy}} & Noun \#1 & 7,539 \\
\multirow{4}{*}{} & Noun \#2 & 1,944 \\
\multirow{4}{*}{} & Verb \#1 & 1,765 \\
\multirow{4}{*}{} & Verb \#2 & 304 \\
\hdashline[2.5pt/2.5pt]
\multirow{3}{*}{{\it Antonymy}} & \#1 & 91 \\
\multirow{3}{*}{} & \#2 & 574 \\
\multirow{3}{*}{} & \#3 & 2,780 \\
\hdashline[2.5pt/2.5pt]
\multirow{3}{*}{{\it Morphosemantic Links}} & Agent & 829 \\
\multirow{3}{*}{} & Instrument & 348 \\
\multirow{3}{*}{} & Result & 788 \\
\hline
{\bf Total} & {\bf --} & {\bf 16,972} \\
\hline
\end{tabular}
}
\caption{\label{table:CQs} Creation of problems on the basis of QPs}
\end{table}

The competency of \SUMO{}-based ontologies can be automatically evaluated by using the framework described in \'Alvez et al. \shortcite{ALR19} and the resources mentioned above. This framework is based on the use of {\it competency questions} (CQs) or {\it problems} \cite{GrF95} derived from the knowledge in \WORDNET{} and its mapping to \SUMO{} by means of several predefined {\it question patterns}. In this paper, we have considered the following QPs:
\begin{itemize}
\item The four QPs based on {\it hyponymy} ---2 QPs for nouns and 2 QPs for verbs--- and the three QPs based on {\it antonymy} introduced in \'Alvez et al. \shortcite{ALR19}.
\item The three QPs based on the {\it Morphosemantic Links} {\it agent}, {\it instrument} and {\it result} introduced in \'Alvez et al. \shortcite{ALR17}.
\end{itemize}
In Table \ref{table:CQs}, we report on the number of CQs/problems that results from each QP.

\begin{figure}[ht]
%\begin{tcolorbox}[standard jigsaw,opacityback=0]
\centering
\begin{tikzpicture}[>=triangle 60]
\matrix[matrix of math nodes,column sep={-3pt},row sep={50pt,between origins},nodes={asymmetrical rectangle}] (s)
{
|[name=hyper]| \langle \synsetTikZ{breathing}{1}{n} \rangle & : & |[name=hyperMappingClass]| [ \equivalenceMappingTikZOfConcept{\SUMOClassTikZ{Breathing}} ] \\
|[name=hypo]| \langle \synsetTikZ{smoking}{1}{n} \rangle & : & |[name=hypoMappingClass]| [ \equivalenceMappingTikZOfConcept{\SUMOClassTikZ{Smoking}} ] \\
};
\draw[-latex] (hypo) -- node[left] {\(\langle hyp \rangle\)} (hyper);
\end{tikzpicture}
%\end{tcolorbox}
\caption{Example for Noun \#2: \synset{smoking}{1}{n} and \synset{breathing}{1}{n}}
\label{fig:CQExample}
\end{figure}
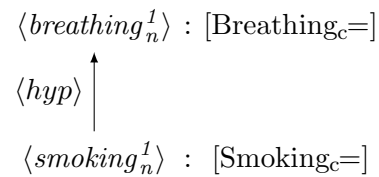

For example, the second QP based on {\it hyponymy} focuses on pairs of hyponym synsets \pair{}{hypo}{hyper} such that the hyponym $hypo$ is connected to \SUMO{} using \equivalenceMappingRelation{}. In those cases, the semantics of $hypo$ is equivalent to the semantics of the \SUMO{} statement that results from its mapping information. Further, the semantics of $hyper$ is more general than the semantics of $hypo$. Consequently, we can state that the set of \SUMO{} instances related to $hyper$ is a superset of the set of \SUMO{} instances connected to $hypo$. %In particular, the noun synset \synset{smoking}{1}{n} (``{\it the act of smoking tobacco or other substances}''), which is connected to \equivalenceMapping{\SUMOClass{Smoking}}, is hyponym of \synset{breathing}{1}{n} (``{\it the bodily process of inhalation and exhalation; the process of taking in oxygen from inhaled air and releasing carbon dioxide by exhalation}''), which is connected to \equivalenceMapping{Breathing} (see Figure \ref{fig:CQExample}). The corresponding \SUMO{} statements by using the same variable are:
%\begin{flalign}
%\doubletab & ( \predicate{instance} \; \variable{X} \; \constant{Smoking} ) & \label{subCQ:smoking} \\
% & ( \predicate{instance} \; \variable{X} \; \constant{Breathing} ) & \label{subCQ:breathing}
%\end{flalign}
%By the instantiation of second QP based on hyponymy using the above two statements, we obtain the following CQ that states that every instance of \SUMOClass{Smoking} is also instance of \SUMOClass{Breathing}:
In particular, the noun synset \synset{smoking}{1}{n} (``{\it the act of smoking tobacco or other substances}'') is hyponym of \synset{breathing}{1}{n} (``{\it the bodily process of inhalation and exhalation; the process of taking in oxygen from inhaled air and releasing carbon dioxide by exhalation}'') (see Figure \ref{fig:CQExample}). By the instantiation of the second QP based on hyponymy using statements (\ref{subCQ:smoking}-\ref{subCQ:breathing}) which result from their mapping information, the following CQ that states that every instance of \SUMOClass{Smoking} is also instance of \SUMOClass{Breathing} can be obtained:
\begin{flalign}
% ;; Goal key: instrumentRelation0287
% ;;	Role: instrument(01624169-v,03699975-n)
% ;;		01624169-v --> [ (machine,1)]
% ;;		03699975-n --> [ (machine,1)]
% ;;		Mapping of the 1st argument: [ (Making,subsumption)]
% ;;		Mapping of the 2nd argument: [ (Machine,equivalence)]
% ;;		Action: instrumentRelation
% ;;	Role: instrument(01623967-v,03699975-n)
% ;;		01623967-v --> [ (machine,2)]
% ;;		03699975-n --> [ (machine,1)]
% ;;		Mapping of the 1st argument: [ (Making,subsumption)]
% ;;		Mapping of the 2nd argument: [ (Machine,equivalence)]
% ;;		Action: instrumentRelation
%
% (forall (?Y)
% 	(=> 
% 		($instance ?Y Machine)
% 		(exists (?X)
% 			(and 
% 				($instance ?X Making)
% 				(instrument ?X ?Y)
% 			)
% 		)
% 	)
% )
\minitab & ( \connective{forall} \; ( \variable{X} ) & \label{CQ:SmokingBreathing} \\
 & \tab ( \connective{=>} & \nonumber \\
 & \tab \tab ( \predicate{instance} \; \variable{X} \; \constant{Smoking} ) & \nonumber \\
 & \tab \tab ( \predicate{instance} \; \variable{X} \; \constant{Breathing} ) ) ) & \nonumber
\end{flalign}

Given a set of CQs and an ontology, the evaluation framework proposes to perform two dual tests using FOL ATPs for each CQ: the first test is to check whether, as expected, the conjecture stated by the CQ is entailed by the ontology ({\it truth-test}); the second one is to check its complementary ({\it falsity-test}). If ATPs find a proof for either the truth- or the falsity-test, then the CQ is classified as {\it solved} (or {\it resolved}). In particular, the CQ is {\it passing}/{\it non-passing} if ATPs find a proof for the truth-test/falsity-test. Otherwise (that is, if no proof is found), the CQ is classified as {\it unresolved} or {\it unknown}. In this last case, we do not know whether (a) the conjectures are not entailed by the ontology or (b) although (some of) the conjectures are entailed, ATPs have not been able to find the proof within the provided execution-time and memory limits.

\section{Detailed Analysis of the Experimental Results} \label{section:Analysis}

\begin{table*}[!t]
\centering
\resizebox{\textwidth}{!}{
\begin{tabular}{l;{2.5pt/2.5pt}r;{2.5pt/2.5pt}r;{1.0pt/2.5pt}rr;{1.0pt/2.5pt}rr;{2.5pt/2.5pt}r;{1.0pt/2.5pt}rr;{1.0pt/2.5pt}rr;{2.5pt/2.5pt}r;{1.0pt/2.5pt}rr;{2.5pt/2.5pt}rr;{1.0pt/2.5pt}rrr}

\hline

\multicolumn{1}{c;{2.5pt/2.5pt}}{\multirow{3}{*}{{\bf Question Pattern}}} & \multicolumn{1}{c;{2.5pt/2.5pt}}{\multirow{3}{*}{{\bf \#}}} & \multicolumn{5}{c;{2.5pt/2.5pt}}{\multirow{2}{*}{{\bf Entailed}}} & \multicolumn{5}{c;{2.5pt/2.5pt}}{\multirow{2}{*}{{\bf Incompatible}}} & \multicolumn{3}{c;{2.5pt/2.5pt}}{\multirow{2}{*}{{\bf Unsolved}}} & \multicolumn{5}{c}{{\bf Total}} \\
\multicolumn{1}{c;{2.5pt/2.5pt}}{\multirow{3}{*}{}} & \multicolumn{1}{c;{2.5pt/2.5pt}}{\multirow{3}{*}{}} & \multicolumn{5}{c;{2.5pt/2.5pt}}{\multirow{2}{*}{}} & \multicolumn{5}{c;{2.5pt/2.5pt}}{\multirow{2}{*}{}} & \multicolumn{3}{c;{2.5pt/2.5pt}}{\multirow{2}{*}{}} & \multicolumn{2}{c}{} & \multicolumn{2}{c}{S} & \\
\multicolumn{1}{c;{2.5pt/2.5pt}}{\multirow{3}{*}{}} & \multicolumn{1}{c;{2.5pt/2.5pt}}{\multirow{3}{*}{}} & \multicolumn{1}{c}{S} & \multicolumn{1}{c}{CM} & \multicolumn{1}{c}{IM} & \multicolumn{1}{c}{CK} & \multicolumn{1}{c;{2.5pt/2.5pt}}{IK} & \multicolumn{1}{c}{S} & \multicolumn{1}{c}{CM} & \multicolumn{1}{c}{IM} & \multicolumn{1}{c}{CK} & \multicolumn{1}{c;{2.5pt/2.5pt}}{IK} & \multicolumn{1}{c}{U} & \multicolumn{1}{c}{CM} & \multicolumn{1}{c;{2.5pt/2.5pt}}{IM} & \multicolumn{1}{c}{CM} & \multicolumn{1}{c}{IM} & \multicolumn{1}{c}{CK} & \multicolumn{1}{c}{IK} & \multicolumn{1}{c}{U} \\

\hline

Noun \#1 (7,539) & 80 & 39 & 33 (5) & 6 & 39 & 0 & 15 & 7 (0) & 8 & 15 & 0 & 26 & 19 (0) & 7 & 59 (5) & 21 & 54 & 0 & 26 \\
Noun \#2 (1,944) & 15 & 9 & 9 (5) & 0 & 9 & 0 & 2 & 2 (2) & 0 & 2 & 0 & 4 & 3 (2) & 1 & 14 (9) & 1 & 11 & 0 & 4 \\
Verb \#1 (1,765) & 13 & 5 & 3 (1) & 2 & 5 & 0 & 0 & 0 (0) & 0 & 0 & 0 & 8 & 6 (0) & 2 & 9 (1) & 4 & 5 & 0 & 8 \\
Verb \#2 (304) & 2 & 0 & 0 (0) & 0 & 0 & 0 & 0 & 0 (0) & 0 & 0 & 0 & 2 & 2 (1) & 0 & 2 (1) & 0 & 0 & 0 & 2 \\
Antonym \#1 (91) & 1 & 0 & 0 (0) & 0 & 0 & 0 & 0 & 0 (0) & 0 & 0 & 0 & 1 & 0 (0) & 1 & 0 (0) & 1 & 0 & 0 & 1 \\
Antonym \#2 (584) & 6 & 1 & 0 (0) & 1 & 1 & 0 & 0 & 0 (0) & 0 & 0 & 0 & 5 & 3 (1) & 2 & 3 (1) & 3 & 1 & 0 & 5 \\
Antonym \#3 (2,780) & 33 & 9 & 4 (0) & 5 & 9 & 0 & 0 & 0 (0) & 0 & 0 & 0 & 24 & 7 (0) & 17 & 11 (0) & 22 & 9 & 0 & 24 \\
Agent (829) & 5 & 1 & 1 (0) & 0 & 1 & 0 & 0 & 0 (0) & 0 & 0 & 0 & 4 & 3 (1) & 1 & 4 (1) & 1 & 1 & 0 & 4 \\
Instrument (348) & 2 & 0 & 0 (0) & 0 & 0 & 0 & 0 & 0 (0) & 0 & 0 & 0 & 2 & 2 (2) & 0 & 2 (2) & 0 & 0 & 0 & 2 \\
Result (788) & 12 & 1 & 1 (0) & 0 & 1 & 0 & 0 & 0 (0) & 0 & 0 & 0 & 11 & 6 (4) & 5 & 7 (4) & 5 & 1 & 0 & 11 \\

\hline

{\bf Total problems (16,972)} & {\bf 169} & {\bf 65} & {\bf 51 (11)} & {\bf 14} & {\bf 65} & {\bf 0} & {\bf 17} & {\bf 9 (2)} & {\bf 8} & {\bf 17} & {\bf 0} & {\bf 87} & {\bf 51 (11)} & {\bf 36} & {\bf 111 (24)} & {\bf 58} & {\bf 82} & {\bf 0} & {\bf 87} \\ 
\hline
\end{tabular}
}
\caption{\label{table:Analysis} Detailed analysis of problems}
\end{table*}

In this section, we report on a detailed and manual analysis of the experimental results obtained from a small number of the CQs described in Section \ref{section:Evaluation}.

%For this purpose, we focus on the experimentation using \FOLSUMO{} reported in \'Alvez et al. \shortcite{ALR17,ALR19}, which has been performed in a Intel\textregistered~Xeon\textregistered~CPU E5-2640v3@2.60GHz with 2GB of RAM memory per processor by setting an execution-time limit of 300 seconds and a memory limit of 2GB per test. Some of the most successful ATPs in {\it The CADE ATP System Competition} (CASC) \cite{Sut16} during the last years have used: Vampire\footnote{Using the following parameters: \tt{--proof tptp --output\_axiom\_names on --mode casc -t 600 -m 2048}.} (v2.6, v3.0, v4.1 and v4.2.2) \cite{KoV13} and E\footnote{Using the following parameters: \tt{--auto --proof-object -s --cpu-limit=600 --memory-limit=2048}.} (v2.1) \cite{Sch02}.

From this experimentation, we have randomly selected a sample of 169 problems (1\% of the total) following a uniform distribution and analysed the results obtained for those problems by focusing on two questions: 1) we analyse the quality of mapping of the involved synsets and 2) we analyse the knowledge required for solving the problems. 

Regarding the quality of the mapping (first question), we classify the mapping of synsets as either {\it correct} or {\it incorrect} according to the following criteria: a mapping is classified as {\it correct} if the semantics associated with the \SUMO{} concept and with the synset are compatible, and it is classified as {\it incorrect} otherwise. For example, both the verb synset \synset{machine}{1}{v} and the adjective synset \synset{homemade}{1}{a} are connected to \subsumptionMapping{\SUMOClass{Making}}, where the semantics of the \SUMO{} class \SUMOClass{Making} is {\it ``The subclass of \SUMOClass{Creation} in which an individual \SUMOClass{Artifact} or a type of \SUMOClass{Artifact} is made''}. Since the semantics of the verb synset \synset{machine}{1}{v} is {\it ``Turn, shape, mold, or otherwise finish by machinery''}, we classify the mapping of 
\synset{machine}{1}{v} as {\it correct}. On the contrary, the semantics of the adjective synset \synset{homemade}{1}{a} is {\it ``made or produced in the home or by yourself''}. Thus, we classify the mapping of \synset{homemade}{1}{a} as {\it incorrect}. 

In addition, synsets with a correct mapping are classified as either {\it correct and precise} or {\it only correct}: a correct mapping is also considered as {\it correct and precise} if the semantics of the synset and the \SUMO{} concept are equivalent, and it is classified as {\it only correct} (that is, correct but not precise) if the semantics of the \SUMO{} concept is more general than the semantics of the synset. For example, the mapping of \synset{machine}{1}{v} to \SUMOClass{Making} is classified as {\it only correct} since the semantics of \SUMOClass{Making} is more general than the semantics of \synset{machine}{1}{v}. By contrast, the mapping of the noun synset \synset{machine}{1}{n} to \equivalenceMapping{\SUMOClass{Machine}} is classified as {\it correct and precise} since the semantics of \synset{machine}{1}{n} is {\it ``Any mechanical or electrical device that transmits or modifies energy to perform or assist in the performance of human tasks''} and the semantics of \SUMOClass{Machine} is {\it ``\SUMOClass{Machine}'s are \SUMOClass{Device}'s that that have a well-defined resource and result and that automatically convert the resource into the result''}.

Regarding the required knowledge (second questions), we distinguish three cases:
\begin{itemize}
\item If the problem is solved, then we classify the knowledge in the proof provided by ATPs as either {\it correct} or {\it incorrect} depending on whether it matches our world knowledge or not.
\item If the problem is unsolved and the mapping of the two involved synsets is correct, then we manually check whether the problem can be entailed by the knowledge in the ontology.
\item If the problem is unsolved and the mapping of some of the involved synsets is incorrect, then the knowledge in the problem does not match our world knowledge and, consequently, it is not subject of classification.
\end{itemize}
It is worth noting that, in the case of unsolved problems such that the required knowledge is classified as existing, ATPs cannot find a proof for its truth- or falsity-test because of the lack of time or memory resources.

In Table \ref{table:Analysis} we summarise some figures of our detailed analysis, where problems are organised according to their QP. The name of the QP and the number of resulting CQs is given in the first column (Question Pattern column) and the remaining columns are grouped into five main parts. In the first part (\#, one column), we provide the number of problems of each category that have been randomly chosen. In the second and third parts (Entailed and Incompatible, five columns each), we provide the result of our quality analysis for the solved problems that have been classified as entailed (its truth-test has been proved) and incompatible (its falsity-test has been proved) respectively. Concretely we show:
\begin{itemize}
\item The number of solved problems (S column).
\item The number of solved problems with a correct (CM column) and incorrect mapping (IM column). Additionally, in the CM column we provide the number of solved problems with a correct and precise mapping between brackets.
\item The number of solved problems that have been proved on the basis of correct (CK column) and incorrect knowledge (IK column).
\end{itemize}
In the fourth part (Unsolved, three columns), we provide the result of our analysis for the unsolved problems:
\begin{itemize}
\item The number of unsolved problems (U column).
\item The number of solved problems with a correct (CM column) and incorrect mapping (IM column). As before, in the CM column we provide the number of solved problems with a correct and precise mapping between brackets.
\end{itemize}
Finally, in the last part (Total, five columns) we summarise the result of our analysis:
\begin{itemize}
\item The number of problems with a correct (correct and precise between brackets) and incorrect mapping (CM and IM columns).
\item The number of solved problems (S columns) that have been proved on the basis of correct (CK column) and incorrect knowledge (IK column).
\item The number of unsolved problems (U column).
\end{itemize}

In total, the synsets in 111 problems (66~\%) are decided to be correctly connected to \SUMO{} and, among them, the synsets in 24 problems (14~\%) are decided to be precisely connected. Thus, some of the synsets are not correctly connected to \SUMO{} in 58 problems (34~\%). Further, 82 problems (49~\%) are solved and the knowledge of the ontology that is used in the proofs reported by ATPs is decided to be correct (100~\%) according to our world knowledge. Among solved problems, 65 problems (79~\%) are classified as entailed and 17 problems (21~\%) are classified as incompatible. By manually analysing incompatible problems, we have discovered that the knowledge of \WORDNET{} and \SUMO{} related to all the problems with a correct mapping is not well-aligned. Thus, we can conclude that this reasoning framework also enables the correction of the alignment between \WORDNET{} and \SUMO{}.
%9	nounHyponymPattern2	11439690-n	[ (cloud,1)]	any collection of particles (e.g., smoke or dust) or gases that is visible	[ (Cloud,class,equivalence)]
%		11419404-n	[ (physical_phenomenon,1)]	a natural phenomenon involving the physical properties of matter and energy	NaturalProcess,class,subsumption
For example, \synset{cloud}{1}{n} (``{\it any collection of particles (e.g., smoke or dust) or gases that is visible}'') is hyponym of \synset{physical\_phenomenon}{1}{n} (``{\it a natural phenomenon involving the physical properties of matter and energy}'') in \WORDNET{}. However, \synset{cloud}{1}{n} and \synset{physical\_phenomenon}{1}{n} are respectively connected to \equivalenceMapping{\SUMOClass{Cloud}} and \equivalenceMapping{\SUMOClass{NaturalProcess}}, which are inferred to be disjoint classes in \FOLSUMO{}. 
%Process and Object are disjoint

Further, the mapping of the involved synsets is classified as correct in 51 of 65 entailed problems (78~\%), while only 14 problems (22~\%) are classified as entailed with an incorrect mapping. By contrast, the percentage of problems with an incorrect mapping is much higher among incompatible and unsolved problems: 42~\% (8 of 17 entailed problems) and 41~\% (36 of 87 unsolved problems) respectively. This is especially the case of the problems from the antonym categories: 26 of 40 antonym problems (65~\%) have an incorrect mapping. This fact reveals the poor quality of the mapping of \SUMO{} to \WORDNET{} adjectives. Finally, we have manually checked that 45 of the 51 unsolved problems with a correct mapping (88~\%) cannot be entailed by the knowledge in \SUMO{}, which sets an upper bound on the number of problems that can be classified as solved although augmenting the knowledge of the ontology and correcting the mapping and the alignment between \WORDNET{} and \SUMO{}.

Next, we summarise the main conclusions drawn from our detailed analysis:
\begin{itemize}

\item The solutions of all the solved problems (with either correct or incorrect mapping) are based on correct knowledge of the ontology (CK columns). This means that we have not discovered incorrect knowledge in the ontology by inspecting the proofs provided by ATPs. 

\item The mapping of a half third of the problems is classified as incorrect (58 of 169 problems) and, among them, almost a half of the problems belong to the antonym categories (26 of 58 problems). This is mainly due to the poor quality of the mapping of \SUMO{} to \WORDNET{} adjectives because many of them are connected to \SUMO{} processes instead of \SUMO{} attributes. Further, the number of problems with a precise mapping among the problems with a correct mapping is very low (24 of 111 problems). However, this is not surprising due to the large difference between the number of concepts defined in the core of \SUMO{} (around 3,500 concepts) and \WORDNET{} (117,659 synsets).

\item Among incompatible problems, the ones with a correct mapping (9 of 17 problems) enable the detection of misalignments between the knowledge of \WORDNET{} and \SUMO{}.

\item The number of solved problems among the {\it Morphosemantic Links} problems with a correct mapping is very low (only 2 of 13 problems), which reveals that \FOLSUMO{} lacks the required information about processes in \SUMO{}.

\item Most of the unsolved problems with a correct mapping ---45 of 51 problems (88~\%)--- are due to the lack of information in the core of \SUMO{}. However, we have also discovered 6 problems for which either its truth- or falsity-test is entailed by knowledge in the core of \SUMO{} although it cannot be proved by ATPs within the given resources of time and memory. Thus, ATPs are able to solve 82 of 88 the problems (93~\%) that are entailed by the current knowledge of the ontology.
\end{itemize}

\section{Exhaustive Analysis of some Problems} \label{section:SomeProblems}

In this section, we present a detailed analysis of some of the examples that have been reported in Table \ref{table:Analysis}.

%two passing cases and three non-passing cases. 

%three cases the resulting CQs are non-passing. 

%\textcolor{blue}{Itziar: no he puesto las definiciones de las clases de SUMO en todos los ejemplos, pero habr\'ia que ponerlas, no?}

\subsection{Examples of Entailed Problems}

Next, we present two examples among the 65 problems that are classified as entailed. The mapping information is correct in the first example, while it is incorrect in the second one.

%\subsubsection{Case 1: Aligned knowledge} %Nouns denoting Processes
\label{sec:chypo1}
\subsubsection{Case 1: Correct mapping} %Nouns denoting Processes
\label{sec:chypo1}

The first example we present involves the synsets \synset{army}{1}{n} (``{\it a permanent organization of the military land forces of a nation or state}'') and \synset{armed\_service}{1}{n} (``{\it a force that is a branch of the armed forces}''). These synsets are respectively mapped to the \SUMO{} classes \SUMOClass{Army} and \SUMOClass{MilitaryService} by equivalence. 

In \WORDNET{} \synset{army}{1}{n} is hyponym of \synset{armed\_service}{1}{n} and in \SUMO{} \SUMOClass{Army} is subclass of \SUMOClass{MilitaryService}. In this case, the knowledge in both resources and in the mapping is correctly aligned, so we get an entailed problem. In Table \ref{table:Analysis}, we report 51 entailed problems with a correct mapping.

%[ (simian,1)]	an ape or monkey	[ (Primate,class,subsumption)]	YES	YES	PASSING
%[ (primate,2)]	any placental mammal of the order Primates; has good eyesight and flexible hands and feet	Primate,class,equivalence	YES	YES	

%[ (army,1), (ground\_forces,1), (regular\_army,1)]	a permanent organization of the military land forces of a nation or state	[ (Army,class,subsumption)]	YES	YES	PASSING
%[ (armed\_service,1), (military\_service,1), (service,6)]	a force that is a branch of the armed forces	MilitaryService,class,equivalence	YES	YES	
\subsubsection{Case 2: Incorrect mapping} %Nouns denoting Processes
\label{sec:chypo3}

The second example of entailed problem involves the synsets \synset{atmospheric\_electricity}{1}{n} (``{\it electrical discharges in the atmosphere}'') and \synset{electrical\_discharge}{1}{n} (``{\it a discharge of electricity}''). These synsets are respectively mapped to the \SUMO{} classes \SUMOClass{Lightning} and \SUMOClass{Radiating} by subsumption.

These synsets are related by hyponymy-hyperonymy in \WORDNET{} and by subclass in \SUMO{}, as in the previous case. But, the mapping seems misleading for \synset{electrical\_discharge} to \SUMOClass{Radiating}: (``{\it Processes in which some form of electromagnetic radiation, e.g. radio waves, light waves, electrical energy, etc., is given off or absorbed by something else.}"). However, this case is resolved because by chance the knowledge in \WORDNET{} and in the incorrect mapping to \SUMO{} is aligned.

We have discovered 14 entailed problems with an incorrect mapping.

%aunque con errores (mapping), correcto
%[ (atmospheric\_electricity,1)]	electrical discharges in the atmosphere	[ (Lightning,class,subsumption)]	NO	---	PASSING
%[ (electrical\_discharge,1)]	a discharge of electricity	Radiating,class,subsumption	NO	---	

%[ (bank,10)]	a flight maneuver; aircraft tips laterally about its longitudinal axis (especially in turning); "the plane went into a steep bank"	[ (FlyingAircraft,class,subsumption)]	NO	---	PASSING
%[ (airplane\_maneuver,1), (flight\_maneuver,1)]	a maneuver executed by an aircraft	FlyingAircraft,class,subsumption	NO	---	

% (subclass Lightning Radiating)

\subsection{Examples of Incompatible Problems}

Next, we present three examples of problems that are classified as incompatible due to several reasons.

\subsubsection{Case 1: Knowledge misalignment} %Nouns denoting Processes knowledge misalignment
\label{sec:hypo1}

The first example we present involves the \SUMO{} classes \SUMOClass{Smoking} and \SUMOClass{Breathing} and the synsets \synset{smoking}{1}{n} (``{\it the act of smoking tobacco or other substances}'') and \synset{breathing}{1}{n} (``{\it the bodily process of inhalation and exhalation; the process of taking in oxygen from inhaled air and releasing carbon dioxide by exhalation}'') .

The synset \synset{smoking}{1}{n} is hyponym of \synset{breathing}{1}{n} in \WORDNET{}, which are respectively connected to \equivalenceMapping{\SUMOClass{Smoking}} and \equivalenceMapping{\SUMOClass{Breathing}}. These classes are disjoint in \SUMO{}. That is, instances of \SUMOClass{Smoking} cannot be instances of \SUMOClass{Breathing}. So, according to the knowledge in \SUMO{}, it is not possible to breath and smoke at the same time, but, according to \WORDNET{} smoking is a subtype of breathing. In this case we have, therefore, a knowledge misalignment problem: the knowledge in one of the resources contradicts the knowledge in the other one.

% (disjoint AutonomicProcess IntentionalProcess)	
 
% \subsection{Case 2: Hyponymy with Nouns denoting \textcolor{red}{Natural phenomena}}
% \label{sec:hypo2} 

Another example of this type of cases involves the \SUMO{} classes \SUMOClass{Cloud} and \SUMOClass{NaturalProcess} and the synsets \synset{cloud}{1}{n} (``{\it any collection of particles (e.g., smoke or dust) or gases that is visible}'') and \synset{physical\_phenomenon}{1}{n} (``{\it a natural phenomenon involving the physical properties of matter and energy }''), which are mapped to \SUMO{} respectively by equivalence and subsumption.

In \WORDNET{} \synset{cloud}{1}{n} is hypomym of \synset{physical\_phenomenon}{1}{n}, but in \SUMO{} they belong to different hierarchies: \SUMOClass{Cloud} is subclass of \SUMOClass{Substance} and \SUMOClass{NaturalProcess} is subclass of \SUMOClass{Process}, and these classes are disjoint as in the previous example.

From the incompatible problems reported in Table \ref{table:Analysis}, the knowledge is misaligned in 5 problems.

\subsubsection{Case 2: Imprecise mappings} %Antonymy with Verbs
\label{sec:anto1}

The next example involves the \SUMO{} classes \SUMOClass{Transfer} (``{\it Any instance of Translocation where the agent and the patient are not the same thing.}'') and \SUMOClass{Removing} (``{\it The Class of Processes where something is taken away from a location. Note that the thing removed and the location are specified with the CaseRoles patient and origin, respectively.}''). The involved synsets are \synset{fetch}{1}{v} (``{\it go or come after and bring or take back}'') and \synset{carry\_away}{1}{v} (``{\it remove from a certain place, environment, or mental or emotional state; transport into a new location or state}''). \synset{fetch}{1}{n} is mapped to \SUMOClass{Transfer} via equivalence while \synset{carry\_away}{1}{n} is mapped to \SUMOClass{Removing} via subsumption.

\synset{fetch}{1}{v} and \synset{carry\_away}{1}{v} are antonyms in \WORDNET{}, but their corresponding \SUMO{} classes are related via subclass in \SUMO{}: \SUMOClass{Removing} is subclass of \SUMOClass{Transfer}. In our opinion this is a case of imprecise mapping, although correct, the class \SUMOClass{Transfer} is too general for the synset \synset{fetch}{1}{v}.

In Table \ref{table:Analysis}, we report two incompatible problems with a correct but imprecise mapping.

\subsection{Examples of Unresolved Problems} \label{sec:Unknown}
 
Next, we present two examples of problems that are unresolved due to different causes.

 \subsubsection{Case 1: Lack of knowledge}
 \label{sec:Unknown1}
 
 The first example corresponds to the problems that are not solved due to the lack of knowledge in the ontology and involves the synsets \synset{machine}{1}{v} (\subsumptionMapping{\SUMOClass{Making}}) and \synset{machine}{1}{n} (\equivalenceMapping{\SUMOClass{Machine}}). These synsets are related via morphosemantic relation \textPredicate{instrument}. However, there is no similar knowledge encoded in \SUMO{}, so this example remains unresolved.
 
 In Table \ref{table:Analysis}, we report 45 problems with a correct mapping that are unresolved due to the lack of knowledge in the ontology.

% ;; Goal key: instrumentRelation0287
% ;;	Role: instrument(01624169-v,03699975-n) 
% ;;		01624169-v --> [ (machine,1)] turn, shape, mold, or otherwise finish b...
% ;;		03699975-n --> [ (machine,1)] any mechanical or electrical device that...
% ;;		Mapping of the 1st argument: [ (Making,subsumption)]
% ;;		Mapping of the 2nd argument: [ (Machine,equivalence)]
% ;;		Action: instrumentRelation
% ;;	Role: instrument(01623967-v,03699975-n)
% ;;		01623967-v --> [ (machine,2)]
% ;;		03699975-n --> [ (machine,1)]
% ;;		Mapping of the 1st argument: [ (Making,subsumption)]
% ;;		Mapping of the 2nd argument: [ (Machine,equivalence)]
% ;;		Action: instrumentRelation

% (forall (?Y)
% 	(=> 
% 		($instance ?Y Machine)
% 		(exists (?X)
% 			(and 
% 				($instance ?X Making)
% 				(instrument ?X ?Y)
% 			)
% 		)
% 	)
% )
 
% |[name=instrument]| \langle \synset{machine}{1}{v} \rangle & : & |[name=instrumentMappingClass]| [ \subsumptionMappingTikZOfConcept{\SUMOClassTikZ{Making}} ] \\
% |[name=process]| \langle \synsetTikZ{machine}{1}{n} \rangle & : & |[name=processMappingClass]| [ \equivalenceMappingTikZOfConcept{\SUMOClassTikZ{Machine}}

 \subsubsection{Case 2: Lack of resources}
 \label{sec:Unknown2}
 
The second example corresponds to resolvable problems that remain unresolved due to the lack of resources (mainly time) of ATPs. This example involves the synset \synset{male}{3}{a} linked to \subsumptionMapping{\SUMOClass{Male}} and the sysnset \synset{female}{1}{a} linked to \equivalenceMapping{\SUMOClass{Female}} as antomyms. In this case, although all the knowledge is correct the ATPs cannot find the prove for it. 
 
Among the problems reported in Table \ref{table:Analysis}, we have found 6 problems with a correct mapping that can be solved but that remain unresolved due to the lack of resources of ATPs.
 
% antonymPattern2	0320	01477077-s	[ (male,3)]	for or pertaining to or composed of men or boys; "the male lead"; "the male population"	[ (Male,attribute,subsumption)]	YES	NO	UNKNOWN
%ant(01477077-s,01477806-a)		01477806-a	[ (female,1)]	being the sex (of plant or animal) that produces fertilizable gametes (ova) from which offspring develop; "a female heir"; "female holly trees bear the berries"	[ (Female,attribute,equivalence)]	YES	YES	
 
% \section{Events}
% \label{sec:event}

% \section{Meronymy}
% \label{sec:mero}

% wine- grape

\section{Conclusion and Future Works} \label{section:Conclusion}

In this paper we have presented a detailed analysis of a sample of large benchmark of commonsense reasoning problems that has been automatically obtained from \WORDNET{}, \SUMO{} and their mapping. 
%By means of this analysis, we have detected some knowledge misalignments, mapping errors, misleading when interpreting the relations and lack of knowledge and resources. This analysis has led us to a more detailed classification of the problems. 

Based on this analysis, we can detect that although the framework enables the resolution of around 49~\% of the total problems, only 36~\% of the total are resolved for the good reasons: 60 problems resolved with a correct mapping. We have also detected that the mapping requires a general revision and correction: in particular, in the case of adjectives. On the contrary, the knowledge in \SUMO{} involved in the revised proofs seems to be correct according to our commonsense knowledge. Further, the problems classified as incompatible enable the detection of misalignments between \WORDNET{} and \SUMO{}, while the problems classified as unknown can be taken as a source of knowledge for the augmentation of \SUMO{}. Actually, we are planning to develop an automatic procedure for the augmentation of \SUMO{} on the basis of the knowledge in \WORDNET{}. Finally, we have detected some problems that can be solved on the basis of the knowledge of \SUMO{} but that are not solved due to the limitation of resources of ATPs. 

% + El mapping especialmente en adjetivos debería revisarse.
% + El conocimiento de SUMO parece que es correcto.
% + Los casos incompatibles son una buena fuente para descubrir conocimiento no alineado entre WN y SUMO 
% + Los no resueltos son una buena fuente para completar el conocimiento de SUMO
% + WN puede ser una fuente de conocimiento para completar SUMO de forma automática
% + Aun con conocimiento correcto los ATPs puede que no resuelvan
% + CWA vs. OWA?  ...

\section*{Acknowledgments}

This work has been partially funded by the the project DeepReading (RTI2018-096846-B-C21) supported by the Ministry of Science, Innovation and Universities of the Spanish Government, the projects  CROSSTEXT (TIN2015-72646-EXP) and GRAMM (TIN2017-86727-C2-2-R) supported  by the Ministry of  Economy, Industry and Competitiveness of the Spanish Government, the Basque Project LoRea (GIU18/182) and BigKnowledge -- {\it Ayudas Fundaci\'on BBVA a Equipos de Investigaci\'on Cient\'ifica 2018}.

\bibliographystyle{acl}
% you bib file should really go here 
\bibliography{biblio}

\end{document}